\documentclass{article} 
\usepackage{iclr2024_conference,times}

\usepackage{amsmath,amsfonts,bm}

\def\eqref#1{equation~\ref{#1}}

\def\1{\bm{1}}

\DeclareMathAlphabet{\mathsfit}{\encodingdefault}{\sfdefault}{m}{sl}
\SetMathAlphabet{\mathsfit}{bold}{\encodingdefault}{\sfdefault}{bx}{n}

\usepackage{longtable}
\usepackage{pifont}
\usepackage{times}
\usepackage{latexsym}
\usepackage{booktabs}
\usepackage{multirow}
\usepackage{graphicx}
\usepackage{enumitem}
\usepackage{algorithm}
\usepackage{algorithmic}
\usepackage{hyperref}       
\usepackage{url}            
\usepackage{amsfonts}       
\usepackage{nicefrac}       
\usepackage{microtype}      
\usepackage{xcolor}    
\usepackage{graphics, bm, physics}
\usepackage{xspace}
\usepackage{subcaption}
\usepackage{tcolorbox}
\usepackage{amsmath}
\usepackage[T1]{fontenc}
\usepackage[utf8]{inputenc}
\usepackage{inconsolata}

\def\model{XMainframe\xspace}

\title{\textbf{\texttt{\textcolor{blue}{\model}}}: A Large Language Model for \\ Mainframe Modernization}

\author{\textbf{Anh T. V. Dau, Hieu Trung Dao, Anh Tuan Nguyen, Hieu Trung Tran}\\
	\textbf{Phong X. Nguyen, Nghi D. Q. Bui}\thanks{Corresponding author: Nghi D. Q. Bui (\href{mailto:nghibdq@fpt.com}{nghibdq@fpt.com})} \\
	\textbf{~}\\
	FPT Software AI Center, Vietnam\\
	\href{https://github.com/FSoft-AI4Code/XMainframe}{\texttt{https://github.com/FSoft-AI4Code/XMainframe}}
}
\iclrfinalcopy 
\begin{document}

\maketitle

\begin{abstract}
Mainframe operating systems, despite their inception in the 1940s, continue to support critical sectors like finance and government. However, these systems are often viewed as outdated, requiring extensive maintenance and modernization. Addressing this challenge necessitates innovative tools that can understand and interact with legacy codebases. To this end, we introduce \textit{\textbf{\model}}, a state-of-the-art large language model (LLM) specifically designed with knowledge of mainframe legacy systems and COBOL codebases. Our solution involves the creation of an extensive data collection pipeline to produce high-quality training datasets, enhancing \model’s performance in this specialized domain. Additionally, we present \textit{\textbf{MainframeBench}}, a comprehensive benchmark for assessing mainframe knowledge, including multiple-choice questions, question answering, and COBOL code summarization. Our empirical evaluations demonstrate that \model consistently outperforms existing state-of-the-art LLMs across these tasks. Specifically, \model achieves 30\% higher accuracy than DeepSeek-Coder on multiple-choice questions, doubles the BLEU score of Mixtral-Instruct 8x7B on question answering, and scores six times higher than GPT-3.5 on COBOL summarization. Our work highlights the potential of \model to drive significant advancements in managing and modernizing legacy systems, thereby enhancing productivity and saving time for software developers.
\end{abstract}

\section{Introduction}

Large Language Models for code (CodeLLMs) excel in processing and understanding source code across various programming languages such as Python, C++, Java, C\#, Rust, Go, etc., as well as descriptive texts~\cite{qin2023chatgpt,touvron2023llama,roziere2023code,jiang2024mixtral,team2024codegemma,manh2023vault,zheng2024opencodeinterpreter,li2023starcoder,wang2023codet5+,feng2020codebert,wang2021codet5,bui2023codetf}. Their ability to recognize patterns, syntax, and semantics makes them highly effective at tasks such as code completion, bug detection, and generating human-readable explanations. These models can bridge the gap between code and documentation by comprehending and generating natural language descriptions 

\textbf{Mainframe Modernization}: Mainframe software systems are crucial to the daily operations of many of the world’s largest corporations, including numerous Fortune 1000 companies. These systems are used extensively in domains such as banking, finance, and government, where they manage large-scale user bases and applications. Despite their origins in the 1950s, COBOL (Common Business Oriented Language) remains widely used in mainframe applications. It is estimated that over 220 billion lines of COBOL code are currently in use, with 1.5 billion lines written annually \cite{cobolblog}. Additionally, COBOL systems manage USD 3 trillion in commerce daily \cite{cobolblog2}. However, the retirement of many COBOL developers and mainframe experts poses a significant challenge for maintaining and modernizing these systems. In 2014, American Banker reported that banks face difficulties in attracting young tech talent and there is a shortage of professionals with mainframe and COBOL skills \cite{cobolblog3}. This highlights the urgent need for innovative solutions to bridge the gap between legacy COBOL systems and modern technologies, denoted as mainframe modernization. There is recent interest in adapting mainstream CodeLLMs to modernize legacy systems written in aging languages like COBOL into modern languages such as C++ and Java to address the shortage and retirement of COBOL developers and mainframe experts. 

\textbf{Challenges}: Integrating mainstream CodeLLMs into current mainframe systems for modernization presents significant challenges:
\begin{itemize}[leftmargin=*]
	\item \textbf{Limited training on mainframe languages}: Existing CodeLLMs, despite being trained on a vast array of languages (both natural languages and programming languages), are not properly trained on languages that run on mainframes, such as COBOL. The amount of COBOL code available on the Internet is much smaller compared to other languages, resulting in low-quality understanding and reasoning of COBOL code by these models \cite{puri2021codenet}.
	\item \textbf{Lack of proper benchmarks}: There is a lack of proper benchmarks to evaluate the quality of the results provided by the LLMs due to the absence of comprehensive documentation and clear business goals for such systems. This makes it difficult to measure the effectiveness and reliability of CodeLLMs when applied to mainframe modernization tasks.
	\item \textbf{Complexity beyond code generation}: Existing CodeLLMs are trained mostly for code generation, which is also the most popular use case when adapting CodeLLMs into software engineering tasks. However, the nature of mainframe modernization does not prioritize COBOL code generation, as organizations want to modernize or migrate their systems to other languages. As such, CodeLLMs are required to pursue knowledge beyond code generation to effectively modernize such systems.
\end{itemize}

These challenges underscore the need for specialized approaches when applying CodeLLMs to mainframe modernization. To better understand the potential of CodeLLMs in addressing these challenges, it is crucial to examine the critical tasks in mainframe software systems from a business-oriented perspective:

\begin{itemize}[leftmargin=*]
	\item \textbf{Mainframe System Understanding}: Managing the complexity of mainframe systems requires a deep understanding of their operations. System managers must comprehend the reasons, functions, and methods behind these operations. This task is challenging due to the vast size of the systems, lack of design documents, limited human expertise, and the low expressiveness of legacy code. CodeLLMs can assist by providing automated question-answering systems that analyze these systems and provide accurate answers to managers’ inquiries. These systems can synthesize information from vast amounts of code and documentation, making it easier for managers to gain insights into system operations.
	
	\item \textbf{Legacy Code Interpretation}: Developers today face significant challenges when working with code written in outdated and legacy languages. These legacy systems often lack comprehensive documentation, making it difficult to understand the original intent and functionality of the code. Additionally, the original developers may no longer be available, creating a substantial knowledge gap. To address this, AI systems that assist developers in interpreting legacy codebases must be capable of understanding code at a repository-level scale~\cite{phan2024repohyper, zhang2023repocoder, liu2024codexgraph}. Such tools can provide accurate summaries and descriptions of legacy code, enabling developers to work with these systems more effectively. By generating detailed explanations and summaries, CodeLLMs help bridge the knowledge gap, facilitating the interpretation and maintenance of complex legacy code.
	
	\item \textbf{System Maintenance}: Given the business-critical nature of mainframe systems, maintenance and upgrades are frequent and crucial. Developers need to integrate new features into existing systems, but limited knowledge of the existing system’s architecture, codebase, and interfaces can lead to errors, inefficiencies, and longer development times. CodeLLMs can analyze the system and suggest code modifications, providing developers with deeper insights into the system’s structure. This ensures consistency, reduces the likelihood of introducing bugs, and accelerates development processes. CodeLLMs can also predict potential issues and offer solutions, enhancing the overall maintenance process.
	
	\item \textbf{Accurate Assessment of Migrated Modules}: Ensuring the correctness and functionality of modules that have been migrated from COBOL to modern programming languages is crucial. While manual efforts can facilitate this translation, verifying that the migrated code faithfully replicates the behavior of the original COBOL modules is essential. The absence of rigorous assessment mechanisms may lead to errors and system failures. Given the mission-critical nature of mainframe applications in sectors like banking, finance, and government, automated validation and verification tools powered by LLMs are necessary. These tools can compare the original and migrated code, identify discrepancies, and ensure that the new modules meet the required specifications and business logic. Effective assessment minimizes the risk of introducing bugs and ensures a smooth transition, preserving the integrity of the system’s operations.
\end{itemize}

\textbf{Contributions:} As we believe that the current mainstream CodeLLMs do not possess sufficient knowledge to address the challenges of mainframe modernization, we propose \model, a foundation language model for code that is specialized with knowledge in mainframe systems. This model can serve as the foundational knowledge base, offering specific capabilities related to mainframe modernization tasks, such as understanding and summarizing COBOL code better than other models. It also has the capability to reason and answer questions related to mainframe systems more effectively due to the rigorous training process using our specific pipeline to collect data related to mainframes and COBOL. In addition, we introduce \textit{MainframeBench}, a benchmark to evaluate mainframe knowledge for LLMs that includes three subtasks: \textbf{Multiple Choice Questions (MCQ)}, \textbf{Question Answering}, and \textbf{COBOL code summarization}. In our evaluation pipeline, \model significantly outperforms other state-of-the-art CodeLLMs such as DeepSeek-Coder~\cite{guo2024deepseek} and Mixtral-Instruct 8x7B~\cite{jiang2023mistral} on MainframeBench.
In summary, our work makes the following contributions:

\begin{enumerate}[leftmargin=*]
	\item We introduce \model, a state-of-the-art LLMs for mainframe operating systems and COBOL legacy code.
	
	\item \model is built on top of DeepSeek-Coder and is available in two versions:
	\begin{itemize}
		\item \model-base: the foundation model specifically designed for mainframe systems and COBOL codebases.
		\item \model-instruct: the instruction-tuned model for understanding mainframe instructions and COBOL programs.
	\end{itemize}
	
	\item We propose a data collection pipeline within \model to produce high-quality datasets. This pipeline enhances \model’s capabilities to leverage knowledge for understanding this particular domain.
	
	\item We provide \textit{MainframeBench}, a standard benchmark for mainframe knowledge, which includes three subtasks: Multiple Choice Questions, Question Answering, and COBOL code summarization.
	
	\item In our benchmark evaluation, \model outperforms state-of-the-art publicly available LLMs on all three tasks. Specifically, our instruction-tuned \model surpasses DeepSeek-Coder-instruct with a 30\% increase in accuracy on the multiple-choice question set. For question answering, \model achieves a BLEU score of 22.02, which is double that of Mixtral-Instruct 8x7B and five times better than DeepSeek-Coder-instruct 33B. Additionally, the BLEU score of our LLM on the COBOL summarization task is six-fold that of GPT 3.5 and other open code LLMs.
\end{enumerate}

\section{Related Work}
\subsection{Code Large Language Models}

Numerous Code-LLMs have been trained on massive datasets, leading to significant advancements across various coding tasks, including code generation \cite{roziere2023code,touvron2023llama,li2023starcoder, jiang2024mixtral, feng2020codebert}, code summarization \cite{ahmed2022few,DBLP:journals/corr/abs-2102-04664,gao2023code,su2024distilled, to2023better, bui2018hierarchical, nguyen2022hierarchynet}, and program repair \cite{lesstraining2022,wei2023copiloting, xia2023automated, bui2022detect}. These models have also demonstrated unexpected capabilities, such as adapting to different domains through discrete prompting, without requiring parameter modifications. Human-crafted or LLM-generated prompts, which include instructions and relevant context, are used to refine the generation process \cite{luo2023wizardcoder,wang-etal-2023-self-instruct}. Related to instruction tuning is chain-of-thought prompting, where models are encouraged to explain their reasoning when faced with complex problems, increasing the likelihood of correct answers \cite{NEURIPS2022_9d560961}. Recently, several studies have explored multi-agent collaborations, where each agent specializes in a unique task—such as code generation or task planning—to enhance the effectiveness of LLM-based code generation \cite{chen2023agentverse,qian2023communicative,huang2023agentcoder}.

\subsection{LLMs for Domain-Specific Tasks}
While general LLMs are trained to cover a wide range of topics, they are often outperformed by smaller models trained exclusively on domain-specific data in tasks within those domains \cite{wu2023bloomberggpt, pal2024domain, arefeen2024leancontext}. This has led to the development of specialized LLMs in various areas, such as finance \cite{wu2023bloomberggpt,yang2023fingpt}, law \cite{cui2023chatlaw}, health \cite{yang2022large,peng2023study}, and IT operations \cite{guo2023owl}. The success of these models underscores the benefits and necessity of tailoring AI models to specific fields.

In the context of mainframe systems, which are critical yet underrepresented in AI research, there are very few AI models designed to support tasks in this domain. Granite \cite{mishra2024granite} from IBM is the first model developed for this purpose. However, Granite has limitations: it only supports IBM’s Z-system and focuses primarily on documents rather than source code, resulting in suboptimal performance on coding tasks for legacy systems, such as code completion or code summarization of COBOL codebases. Another model, Mainframer from BloopAI \cite{mainframer}, is one of the few models designed to support coding tasks for legacy COBOL systems, achieving good performance in COBOL code completion. However, it is trained solely on a dataset specific to code completion, rendering it nearly ineffective for other tasks like question answering or code summarization. In contrast, our goal is to build a universal model that excels across various tasks in this domain, delivering high performance consistently.

\subsection{Benchmark for COBOL and Mainframe Systems}
Code-related datasets have been developed to facilitate empirical research across various programming languages and address challenges in multiple areas of software engineering \cite{50670,concode,chen2021evaluating,nguyen-etal-2023-vault}. However, low-resource languages like COBOL have received limited attention from the scientific and academic communities, creating a significant barrier to training LLMs for COBOL on a large scale. OpenCBS \cite{lee2022opencbs} is one of the pioneering efforts in this space, leveraging public forums to create a COBOL dataset for defect detection. Another dataset, X-COBOL \cite{ali2023x}, consists of 84 COBOL repositories collected from GitHub. Despite undergoing a data extraction pipeline, this dataset falls short in quality because the authors relied on GitHub stars for filtering repositories, which is not a reliable metric for determining repository quality \cite{borges2018s}. More recently, BloopAI announced COBOLEval \cite{mainframer}, a benchmark designed to evaluate legacy code completion tasks. It consists of 146 coding problems converted into COBOL from the HumanEval benchmark \cite{chen2021evaluating}, originally a Python dataset. However, this approach is unrealistic, as COBOL is primarily used in business and finance systems, not for solving general programming challenges. To the best of our knowledge, there is no dataset that comprehensively covers diverse tasks related to the COBOL language and legacy systems.

\section{Data Construction}



Data quality plays a vital role in training large language models, which directly affects their performance \cite{shi2022we,dau-etal-2024-docchecker}. Although training datasets for language and code are popular and high-quality \cite{laurenccon2022bigscience, nguyen-etal-2023-vault}, finding a dataset to support various tasks within mainframe system understanding and legacy coding is challenging. 
To support fine-tuning \model, we build from scratch our own dataset specific to this domain. 
In the following sections, we introduce our Mainframe-Training Dataset \ref{sec:Mainframe_training_dataset} and Mainframe-Instruct Dataset. \ref{sec:mainframe_instruct_dataset}, which are used for training and instruction tuning, respectively.

\subsection{Dataset for Pretraining}

\label{sec:Mainframe_training_dataset}
This section details the data extraction process for training \model. We utilized two different sources: using the GitHub API to collect COBOL projects hosted on GitHub and gathering online document data relevant to mainframes.

\begin{figure}[]
\centering
\includegraphics[width=\textwidth]{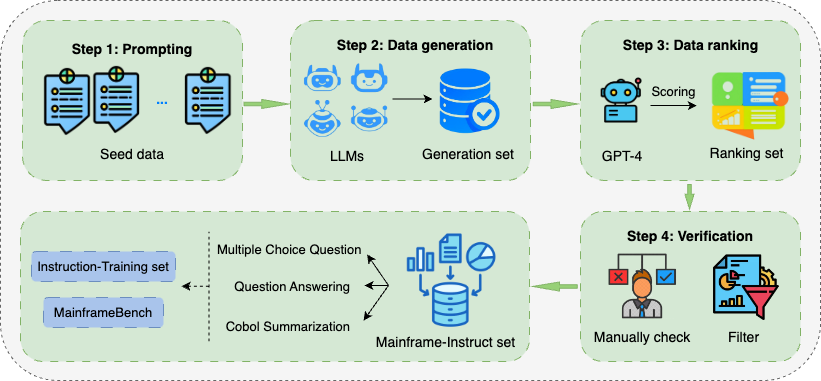}
\caption{Data Augmentation Pipeline.}
\label{fig:mainframe_instruct_pipeline}
\end{figure}

We initially retrieved all GitHub repositories containing COBOL and Mainframe system code, amassing approximately 4GB of data. To ensure high-quality training samples, we removed overly short repositories and files, eliminated alphanumeric character fractions, binary data, JSON, XML data, and node modules, resulting in 40,960 COBOL files. We further refined our dataset using MinHash and Locality Sensitive Hashing (LSH) to detect and remove near-duplicates [citation]. This process involved document shingling and fingerprinting, using locality-sensitive hashing to group similar documents, detecting actual duplicates, and removing them. The final COBOL dataset consists of 33,561 files, encompassing 228 million tokens in 8 million Lines of Code (LoCs). For Mainframe documents, we extracted data from public books and websites related to Mainframe and COBOL, ensuring minimal noise and maximum data cleanliness. We extracted main content from HTML pages and eliminated unnecessary parts using specific tags, IDs, and keywords, resulting in 14,274 documents containing approximately 8 million tokens.

\begin{figure*}[]
\centering
\includegraphics[width=\textwidth]{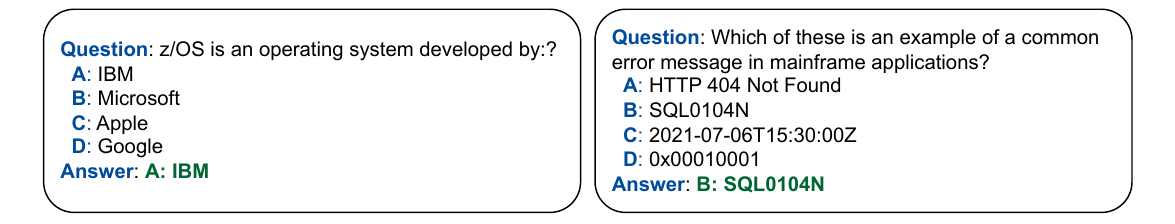}
\caption{Examples for Multiple Choice Question task.}
\label{fig:MC}
\end{figure*}

In total, the training dataset consists of 236 million tokens from documents about the mainframe technology and COBOL constructs. This data collection phase is not only foundational in pretraining LLMs for COBOL and mainframe systems but also a robust groundwork for the model's subsequent instruction fine-tuning, promising significant improvement in its predictive and generative capabilities within this specialized domain.

\begin{table*}[]
\centering
\begin{tabular}{@{}l|ccc@{}}
\toprule
\multicolumn{1}{c|}{\textbf{}} & \multicolumn{1}{c}{\textbf{Train}} & \multicolumn{1}{c}{\textbf{Validation}} & \multicolumn{1}{c}{\textbf{Test}} \\ \midrule
Multiple Choice Questions      & 13.894                                        & 1.544                                           & 1.931                                        \\
Question Answering             & 18.692                                        & 2.078                                           & 2.598                                        \\
COBOL Summarization            & 9.081                                         & 1.010                                           & 2.523                                        \\ \bottomrule
\end{tabular}
\caption{Statistics of the Instruction Dataset. The test set is the MainframeBench benchmark, which is used consistently throughout our evaluation pipeline.}
\label{tab:statistics}
\end{table*}

\subsection{Dataset for Model Instruct}

\label{sec:mainframe_instruct_dataset}
In order to maintain a high-quality synthetic dataset, we employ a pipeline to construct the Instruction dataset, which consists of five distinct phases and is shown in Figure \ref{fig:mainframe_instruct_pipeline}. 

In the initial stage, 300 seed data instances about Mainframe and COBOL are gathered and annotated by our domain experts in the QA formats \cite{zhang2023instruction}. They include knowledge-based question answering, deployment, syntax, COBOL code summarization, and other aspects. To cover various practical scenarios, we design long and short versions with different styles of instruction prompts for each type of question, assisting with our large language model's supervised instruction-tuning process. Then, we use OpenAI GPT-4-turbo to generate more than 200 sub-topics within the Mainframe and COBOL fields. This step is designed to ensure the generated content is firmly rooted in our specific domain. All of them serve as the foundation for data augmentation, enhancing the scale and diversity of our dataset.

Besides using LLMs to solve tasks, recent works have treated LLMs as data generators \cite{ye-etal-2022-zerogen,ye-etal-2023-generating,yu2024large}. With only a few examples, LLMs are able to generate more high-quality data through in-context learning and prompting. The experiments showed that task-specific models trained on generated data can beat the performance of original LLMs while maintaining a low inference cost.
Inspired by the self-instruct approach \cite{ouyang2022training}, we further enrich Mainframe-Instruct from the seed data by harnessing the capabilities of popular LLMs, including OpenAI GPT-3.5-turbo, Mistral-Instruct 7B \cite{jiang2023mistral}, Neural-Chat 7B \cite{Intel}, and Mixtral-Instruct 8x7B model \cite{jiang2024mixtral}, which are trained on numerous languages and achieved high performance on various NLP and software benchmarks. Below are the examples of prompts that we use to generate data from a sub-topic and seed data:

\begin{figure*}[]
\centering
\includegraphics[width=\textwidth]{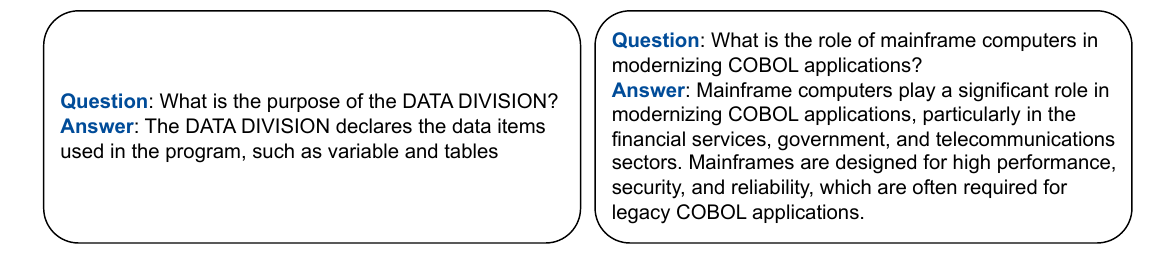}
\caption{Examples for Question Answering task.}
\label{fig:QA}
\end{figure*}

\begin{figure}[]
\centering
\includegraphics[width=0.8\textwidth]{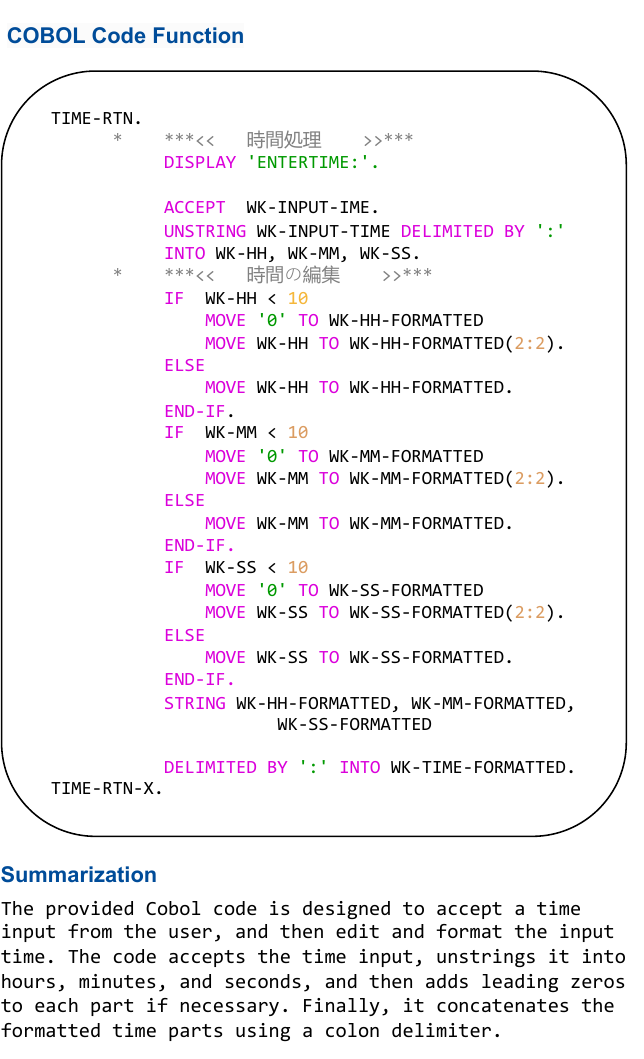}
\caption{Examples for COBOL summarization task.}
\label{fig:sum}
\end{figure}

\begin{tcolorbox}
\textbf{Prompt to generate data from sub-topic:} You have been provided with a Mainframe-related topic, specifically \textit{[sub-topic]}. Your
task is to produce a comprehensive list of question-answer pairs adhering to
the following guidelines:

1. All generated questions must pertain to the specified topic.

2. The question should be detailed.

3. The answers must accurately address the corresponding question,
eliminating unnecessary details while retaining essential information.

4. Format: You are allowed to provide only a list of parsable JSON format
data. Each entry should include:

"question" field containing the question related to the given topic;

"answer" field containing a clear, short, and concise response to the
question

\end{tcolorbox}

\begin{tcolorbox}
  \textbf{Prompt to generate data from seed data:}  I need to build a high-quality COBOL dataset, where each sample is a COBOL paragraph-summary pair.
  
Try to generate multiple examples when possible following this requirement:

1. The generated example must include the COBOL paragraph and the corresponding summary.

2. Format: You are allowed to provide only a list of parsable JSON format data. Each entry should include:

"source" field contains the COBOL paragraph;

"summary" field includes a clear, short, and concise summary of the corresponding COBOL code.

Follow the below example to generate more data:

Example:

"source": \textit{[source]}

"summary":  \textit{[summary]}

\end{tcolorbox}

To ensure a strict standard of data quality, we combine OpenAI GPT-4-turbo with careful manual validation. These steps improve the overall quality of our created data while guaranteeing its integrity and dependability. GPT-4 is utilized as an evaluator to judge model responses, scoring the outputs and ranking responses in a pairwise manner. We design prompts meticulously for this task, making GPT-4 easier to locate and remove any instances of poor-quality data. Finally, the dataset undergoes a rule-based filter and manual inspection by our domain experts. All entries that do not fit our standard are fixed or deleted from the dataset. The prompt used for GPT-4 is presented below:

\begin{tcolorbox}
  \textbf{Quality Prompt:} You are given a list of question-answer pairs that are related to Mainframe Migration and COBOL legacy. By thinking step by step to give the final answer, please help me rate the following pairs according to my requirements.
  
Require:

1. Scoring perspective: whether the question is related to my topic, the answer should be exactly to the corresponding question.

2. Point scale: 10-point scale, from 1-very poor to 10-excellent.
 
3. Format: At the end of your response, you need to add a list of integers, corresponding
the final score for each question-answer pair.

Now, please follow the above requirements to annotate the following data and return your annotated results in a list at the end. 
\end{tcolorbox}

Consequently, the final version of Mainframe-Instruct comprises a total of 53,351 entries and is divided into three tasks: Multiple Choice Questions, Question Answering, and COBOL summarization. Figure \ref{fig:MC}, \ref{fig:QA}, and \ref{fig:sum} are examples corresponding to three tasks. The statistic of this dataset is shown in Table \ref{tab:statistics}. MainframeBench, our benchmark for mainframe knowledge, is the testing set in Mainframe-Instruct Dataset

\section{Overview of \model}
\label{sec:model}
\begin{figure*}[]
\centering
\includegraphics[width=\textwidth]{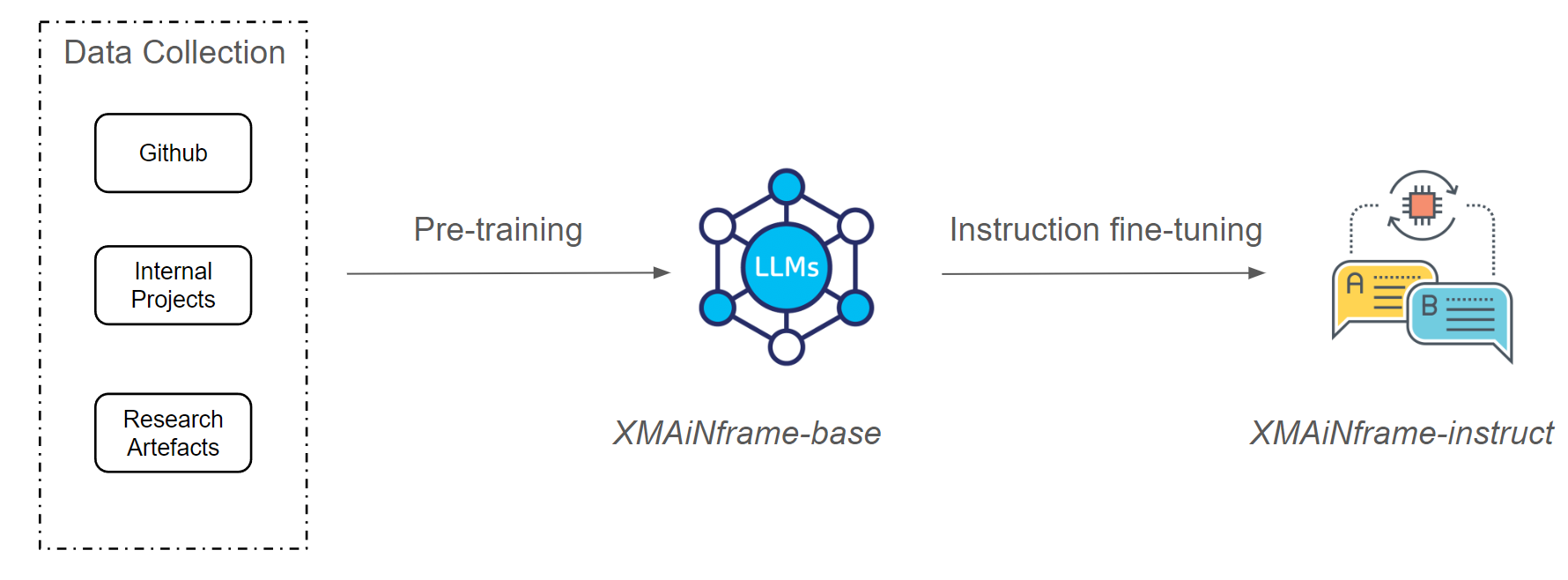}
\caption{Overview of training process.}
\label{fig:training}
\end{figure*}
In this section, we detail the selection of the backbone mode \ref{sec:select}, our training process \ref{sec:training}, and the method to scale up the backbone model \ref{sec:upscaling}.

\subsection{Pretrained Model}
\label{sec:select}

We utilize the pre-trained weights DeepSeek-Coder \cite{guo2024deepseek} as our base model. DeepSeek-Coder’s architecture is based on a decoder-only Transformer and is pre-trained on a high-quality project-level code corpus comprising 87 programming languages. It also incorporates Rotary Position Embedding (RoPE) \cite{su2024roformer}, which extends the context window to 16K, enhancing its ability to handle extended context lengths.

\subsection{Training Details}
\label{sec:training}
We train \model through two stages: pre-training and instruction tuning, as illustrated in Figure \ref{fig:training}. In the first stage, \model-base is initially trained on top of DeepSeek-Coder-base 7B using data from our Mainframe-Training Dataset combined with SlimOrca-Dedup \cite{SlimOrcaDedup}. This combination enriches the model’s mainframe knowledge while retaining its general capabilities. We employ standard autoregressive sequence modeling to predict the next token and utilize the efficient optimization of FlashAttention 2 \cite{dao2022flashattention} for training.
Subsequently, in the second stage, the model undergoes instruction tuning on our Mainframe-Instruct Dataset for three epochs. This tuning process enhances the model’s ability to comprehend and execute natural language instructions, resulting in \model-instruct.


\begin{figure*}[]
\centering
\includegraphics[width=\textwidth]{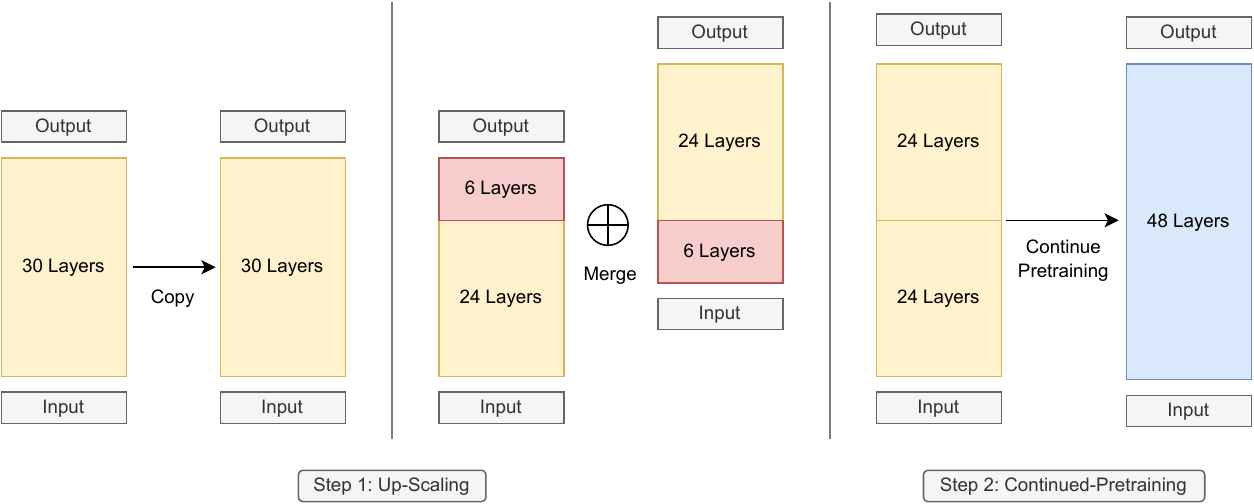}
\caption{Depth up-scaling process.}
\label{fig:up-scaling}
\end{figure*}

\subsection{Model Upscale}
\label{sec:upscaling}

Inspired by \cite{kim2023solar}, we employ the depth up-scaling method to expand the base model without introducing additional modules or dynamic expert selection methods like Mixture of Experts (MoE) \cite{shazeer2017outrageously,komatsuzaki2022sparse}. This approach maintains high efficiency during both training and inference. The depthwise scaling process, illustrated in Figure \ref{fig:up-scaling}, involves two steps: expanding the base model and continuing pretraining. First, the base model, consisting of $n$ layers, is duplicated. We then remove the last $m$ layers from the original model and the first $m$ layers from its duplicate, creating two separate models with $n - m$ layers each. These parts are combined to form a scaled model with $s = 2(n - m)$ layers. For our purposes, we choose $m = 6$. With $n = 30, m = 6, s = 48$, this process is depicted in Step 1 of Figure \ref{fig:up-scaling}. As a result, we scale up DeepSeek-Coder 7B, originally with 30 layers, to a 10.5B model with 48 layers.

Previous experiments have shown that depthwise scaled models initially perform worse than their base counterparts \cite{komatsuzaki2022sparse,kim2023solar}. However, the authors of \cite{kim2023solar} found that the depthwise scaling method isolates the heterogeneity in the scaled model, enabling quick performance recovery. This finding aligns with our experimental results.
Therefore, we continue to train the scaled model on our Mainframe-Training Dataset and fine-tune it on the Mainframe-Instruct Dataset, as shown in Figure \ref{fig:training}, resulting in \model-Instruct 10.5B.
\section{Experiments}
\label{sec:experiments}
\subsection{Experimental Settings}
 We conduct a comparison with other popular LLMs on \textit{MainframeBench}, comprising three subsets: Multiple Choice Questions, Question Answering, and COBOL summarization. 
 Our LLMs are compared with a range of previous state-of-the-art LLMs, including GPT 3.5, GPT-4, Mistral 7B, Mixtral 8x7B, Neural-Chat, DeepSeek Coder 6.7B and 33B. We evaluate these LLMs using zero-shot prompting and fixing the temperature hyperparameter to approximately 0, leading to more exploitation of the model's current knowledge.

\subsection{Metrics}

 \textbf{Metrics for Multiple Choice Question task:} Because it involves the direct model to select a single answer from the provided options (A, B, C, D), it is considered a classification task. We use Accuracy to report the performance of methods on multiple-choice questions. 

 \textbf{Metrics for Question Answering and COBOL Summarization task:} We use various metrics in NLP, including MAP, F1-Score, BertScore, RougeL, Meteor, BLEU-4, as the evaluation metrics for these tasks. These metrics are commonly used to assess the quality and similarity of generated text compared to reference texts.


\subsection{Experiment Results on MainframeBench.}

\subsubsection{Multiple Choice Question}

Table \ref{tab:MC} presents the accuracy scores of various models on a multiple-choice question task. XMAiNframe-Instruct-10.5B stands out with an accuracy score of 77.89\%, which is notably higher than most other models in the comparison. GPT 4 and GPT 3.5 show competitive accuracies of 73.9\% and 74.56\%, respectively. Mixtral-Instruct 8x7B, Mistral-Instruct 7B, and Neural-Chat follow with accuracies ranging from 66.35\% to 69.29\%. 
Although \model-Instruct-7B achieves an accuracy of only 68.5\%, it is 20\% higher than the base model, DeepSeek-Coder-Instruct 7B, and 15\% higher than DeepSeek-Coder-Instruct 33B.
It suggests that \model-Instruct performs exceptionally well on the multiple-choice question task, demonstrating its effectiveness and reliability in this specific domain.
\begin{table}[h]
	\centering
	\begin{tabular}{@{}l|c@{}}
		\toprule
		\textbf{Model} & \textbf{Accuracy (\%)} \\ \midrule
		GPT-4                                & 73.90              \\
		GPT-3.5                              & 74.56             \\
		Mixtral-Instruct 8x7B                & 68.12             \\
		Mistral-Instruct 7B                  & 69.29             \\
		Neural-Chat                          & 66.35             \\
		DeepSeek-Coder-Instruct 6.7B         & 47.49             \\
		DeepSeek-Coder-Instruct 33B          & 53.29             \\
		\model-Instruct 7B                   & 68.57             \\
		\model-Instruct 10.5B                & \textbf{77.89}    \\ \bottomrule
	\end{tabular}
	\vspace{0.5em}
	\caption{Results on Multiple-Choice Question.}
	\label{tab:MC}
\end{table}

\subsubsection{Question Answering}

\model-Instruct demonstrates exceptional effectiveness on the question-answering task, as shown in Table \ref{tab:QA}. With a remarkable MAP of 0.45 and an F1-Score of 0.42, \model-Instruct surpasses all other models in this comparison. Its BLEU-4 score of 20.43 is +9 higher than Mixtral-Instruct 8x7B, which has parameters six times greater than \model-Instruct's. This substantial improvement in scores highlights XMAiNframe-Instruct's ability to provide accurate and contextually relevant answers, making it a highly effective model for question-answering tasks.

\begin{table*}[]
\centering
\begin{tabular}{@{}l|cccccc@{}}
\toprule
\multicolumn{1}{c|}{\textbf{Models}} & \textbf{MAP} & \textbf{F1-Score} & \textbf{BERTScore} & \textbf{RougeL} & \textbf{Meteor} & \textbf{BLEU-4} \\ \midrule
GPT 4                                & 0.12         & 0.19              & 0.88               & 0.18            & 0.34            & 5.71            \\
GPT 3.5                              & 0.14         & 0.22              & 0.89               & 0.21            & 0.38            & 7.36            \\
Mixtral-Instruct 8x7B                 & 0.27         & 0.31              & 0.9                & 0.29            & 0.38            & 11.39           \\
Mistral-Instruct 7B                  & 0.12         & 0.19              & 0.87               & 0.18            & 0.34            & 5.74            \\
Neural-Chat                          & 0.13         & 0.21              & 0.88               & 0.2             & 0.36            & 6.45            \\
DeepSeek-Coder-Instruct 6.7B         & 0.09         & 0.15              & 0.86               & 0.14            & 0.30            & 4.09            \\
DeepSeek-Coder-Instruct 33B          & 0.09         & 0.15              & 0.86               & 0.15            & 0.31            & 4.41            \\
\model-Instruct 7B & \textbf{0.45} & 0.42 & 0.92 & 0.4 &0.42 & 20.43 \\
\model-Instruct 10.5B                         & 0.43        & \textbf{0.42 }             & \textbf{0.92   }            & \textbf{0.4  }          & \textbf{0.42   }         & \textbf{20.93 }         \\ \bottomrule
\end{tabular}
\caption{Results on Question Answering.}
\label{tab:QA}
\end{table*}

\begin{table*}[]
\centering
\begin{tabular}{@{}l|cccc@{}}
\toprule
\multicolumn{1}{c|}{\textbf{Models}} & \textbf{BERTScore} & \textbf{RougeL} & \textbf{Meteor} & \textbf{BLEU-4} \\ \midrule
GPT 4                                & 0.85               & 0.22            & 0.34            & 7.42            \\
GPT 3.5                              & 0.88               & 0.28            & 0.34            & 11.37           \\
Mistral-Instruct 7B                  & 0.85               & 0.12            & 0.15            & 3.61            \\
Neural-Chat                          & 0.88               & 0.27            & 0.34            & 11.07           \\
DeepSeek-Coder-Instruct 6.7B         & 0.85               & 0.22            & 0.32            & 7.72            \\
DeepSeek-Coder-Instruct 33B          & 0.85               & 0.21            & 0.31            & 7.55            \\
\model-Instruct 7B & 0.89 &0.41 & 0.56 & 22.23 \\
\model-Instruct 10.5B               & \textbf{0.96}      & \textbf{0.74}   & \textbf{0.74}   & \textbf{62.58}  \\ \bottomrule
\end{tabular}
\caption{Results on COBOL Code Summarization.}
\label{tab:summarization}
\end{table*}
\subsubsection{COBOL Summarization}

Based on our observations, developers tend to favor concise and comprehensive summary sentences for COBOL code functions over lengthy ones. As a result, the COBOL summarization set in MainframeBench is tailored to reflect this preference. \model-Instruct has the ability to recognize and apply this observation, producing concise and comprehensive summaries. In contrast, other LLMs often generate longer responses that may not align as closely with developers' preferences for summaries. Comparing these models, Table \ref{tab:summarization}
shows that \model-Instruct outperforms the others by a significant margin, achieving notably higher scores on all metrics. Particularly, it achieves a much higher BLEU-4 score, surpassing GPT 3.5 and Neural Chat by approximately six-fold and outperforming GPT-4, DeepSeek-Coder-Instruct 6.7B, and 33B by nine times. This indicates a substantial improvement in the quality and similarity of its generated text compared to the references. While other models show competitive scores, \model-Instruct stands out as the model that excels marginally in this task, showcasing its effectiveness and superior performance on COBOL code understanding.

\section{Conclusion}
In this paper, we present \model, an LLM specifically designed for mainframe operating systems and COBOL legacy codebases. Additionally, we introduce a pipeline to collect and produce high-quality datasets, resulting in a mainframe benchmark that includes three downstream tasks: question answering, multiple choice questions, and COBOL code summarization. Our experiments reveal that \model outperforms existing state-of-the-art LLMs across multiple tasks, demonstrating significant improvements in accuracy and BLEU scores. With its advanced capabilities in knowledge understanding and documentation assistance, \model can boost productivity and transform developers' interaction with and maintenance of mainframe systems and COBOL codebase. Our work not only highlights the benefits of \model but also sets the stage for promoting innovation and efficiency in the management and modernization of legacy systems.

\bibliography{ref}
\bibliographystyle{iclr2024_conference}

\end{document}